\documentclass[letterpaper, 10 pt, conference]{ieeeconf}  

\IEEEoverridecommandlockouts                              

\usepackage{graphicx} 
\usepackage{amsmath} 
\usepackage{amssymb}  

\usepackage{amsthm}  

\theoremstyle{plain}
\newtheorem*{property}{Property}

\newcommand{\suchthat}{{\, : \,}}

\newcommand{\pd}[2]{\frac{\partial #1}{\partial #2}}

\newcommand{\Rbb}{{\mathbb R}}

\newcommand{\dbf}{{\mathbf{d}}}

\newcommand{\ubf}{{\mathbf{u}}}

\newcommand{\Dcal}{{\mathcal{D}}}

\newcommand{\Kcal}{{\mathcal{K}}}
\newcommand{\Lcal}{{\mathcal{L}}}

\newcommand{\Ncal}{{\mathcal{N}}}

\newcommand{\Rcal}{{\mathcal{R}}}

\newcommand{\Ucal}{{\mathcal{U}}}
\newcommand{\Vcal}{{\mathcal{V}}}

\newcommand{\Lcals}{{\mathcal{L}^*}}

\usepackage{mathrsfs}

\newcommand{\Uscr}{{\mathscr{U}}}
\newcommand{\Dscr}{{\mathscr{D}}}

\newcommand{\Vhat}{{\hat{V}}}

\newcommand{\xdot}{{\dot{x}}}

\DeclareMathOperator*{\argmax}{arg\,max}

\makeatletter
\newcommand{\revisionhistory}[1]{%
\@ifundefined{showrevisionhistory}{\relax}{%
{#1}%
}%
}
\makeatother

\title{\LARGE \bf
Modeling Supervisor Safe Sets for Improving Collaboration in Human-Robot Teams
}

\author{David L. McPherson$^{*}$, Dexter R.R. Scobee$^{*}$, Joseph Menke, Allen Y. Yang, S. Shankar Sastry
\thanks{$^{*}$The first two authors contributed equally}
\thanks{This work is supported by the Office of Naval Research under the Embedded Humans MURI (N00014-13-1-0341), NSF Grant CNS-1545126 (VeHICaL), as well as a Philippine-California Advanced Research Institutes (PCARI) grant.}
\thanks{All authors are with the Department of Electrical Engineering and Computer Science,
        University of California, Berkeley
        {\tt\small \{david.mcpherson, dscobee, joemenke, yang, sastry\}@eecs.berkeley.edu }}%
}

\begin{document}

\maketitle
\thispagestyle{empty}
\pagestyle{empty}

\begin{abstract}
When a human supervisor collaborates with a team of robots, the human's attention is divided, and cognitive resources are at a premium.
We aim to optimize the distribution of these resources and the flow of attention.
To this end, we propose the model of an idealized supervisor to describe human behavior.
Such a supervisor employs a potentially inaccurate internal model of the the robots' dynamics to judge safety.
We represent these safety judgements by constructing a \emph{safe~set} from this internal model using reachability theory.
When a robot leaves this safe set, the idealized supervisor will intervene to assist, regardless of whether or not the robot remains objectively safe.
False positives, where a human supervisor incorrectly judges a robot to be in danger, needlessly consume supervisor attention.
In this work, we propose a method that decreases false positives by learning the supervisor's safe set and using that information to govern robot behavior.
We prove that robots behaving according to our approach will reduce the occurrence of false positives for our idealized supervisor model.
Furthermore, we empirically validate our approach with a user study that demonstrates a significant ($p = 0.0328$) reduction in false positives for our method compared to a baseline safety controller.

\end{abstract}
\section{Introduction and Background}
\label{sec:introduction}
As automation becomes more pervasive throughout society, humans will increasingly find themselves interacting with autonomous and semi-autonomous systems.
These interactions have the potential to multiply the productivity of humans workers, since it will become possible for a single human to supervise the behavior of multiple robotic agents.
For example, a single human driver could manage a fleet of self-driving delivery robots, but the driver would only take full control for the ``last mile,'' guiding the robots to precisely deposit packages in environments where autonomous navigation may not be reliable.
Human experts regularly serve as failsafe supervisors on factory assembly floors staffed with robotic arms \cite{hwang1984integration}.
Air traffic controllers soon will have to manage completely autonomous drones flying through their airspace alongside existing traditional mixed-autonomy planes and their auto-pilots \cite{tomlin1999atm}.

\begin{figure}[t]
  \centering
  \includegraphics[width=\columnwidth]{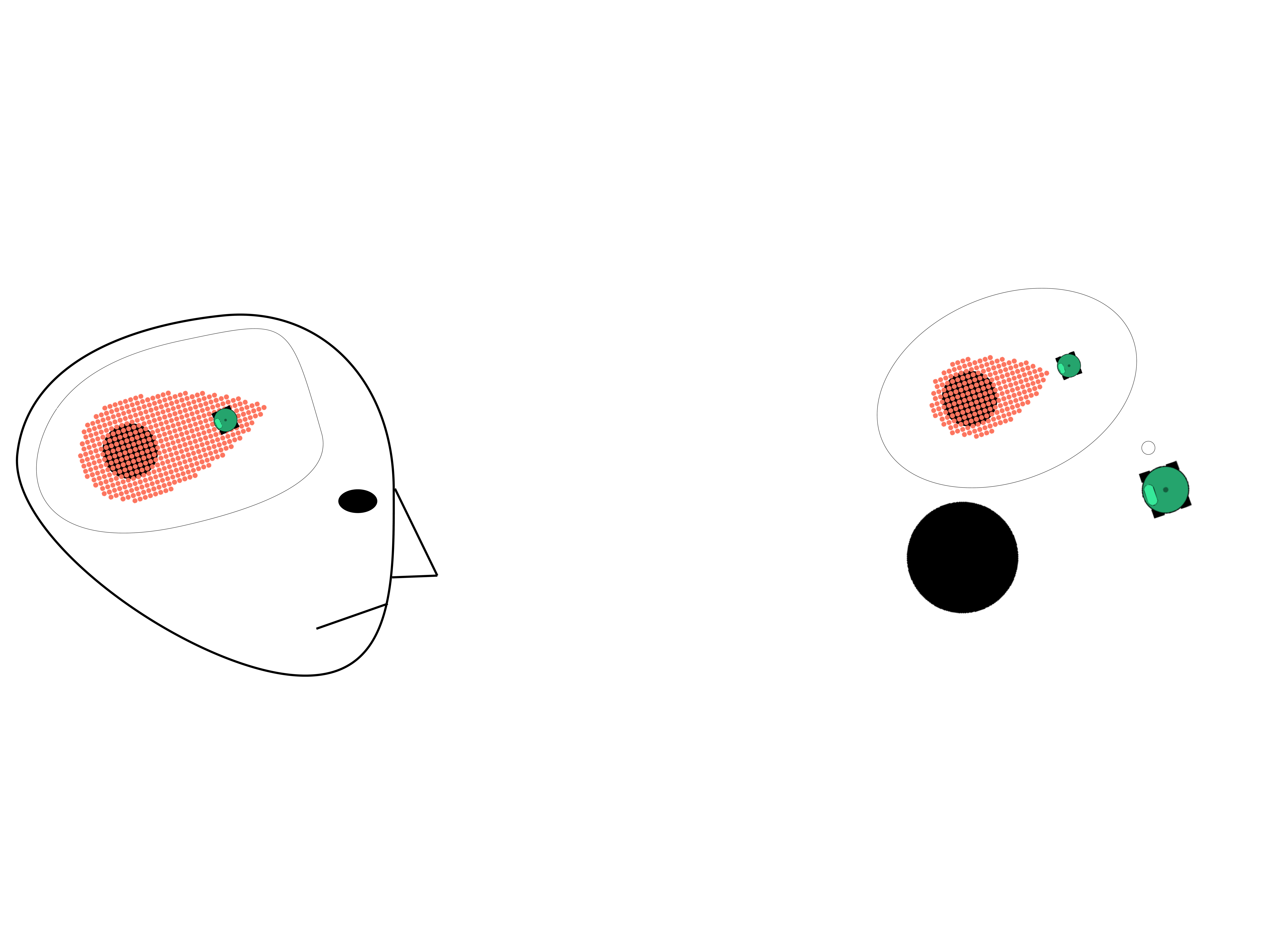}
  \hrule
  \includegraphics[width=\columnwidth]{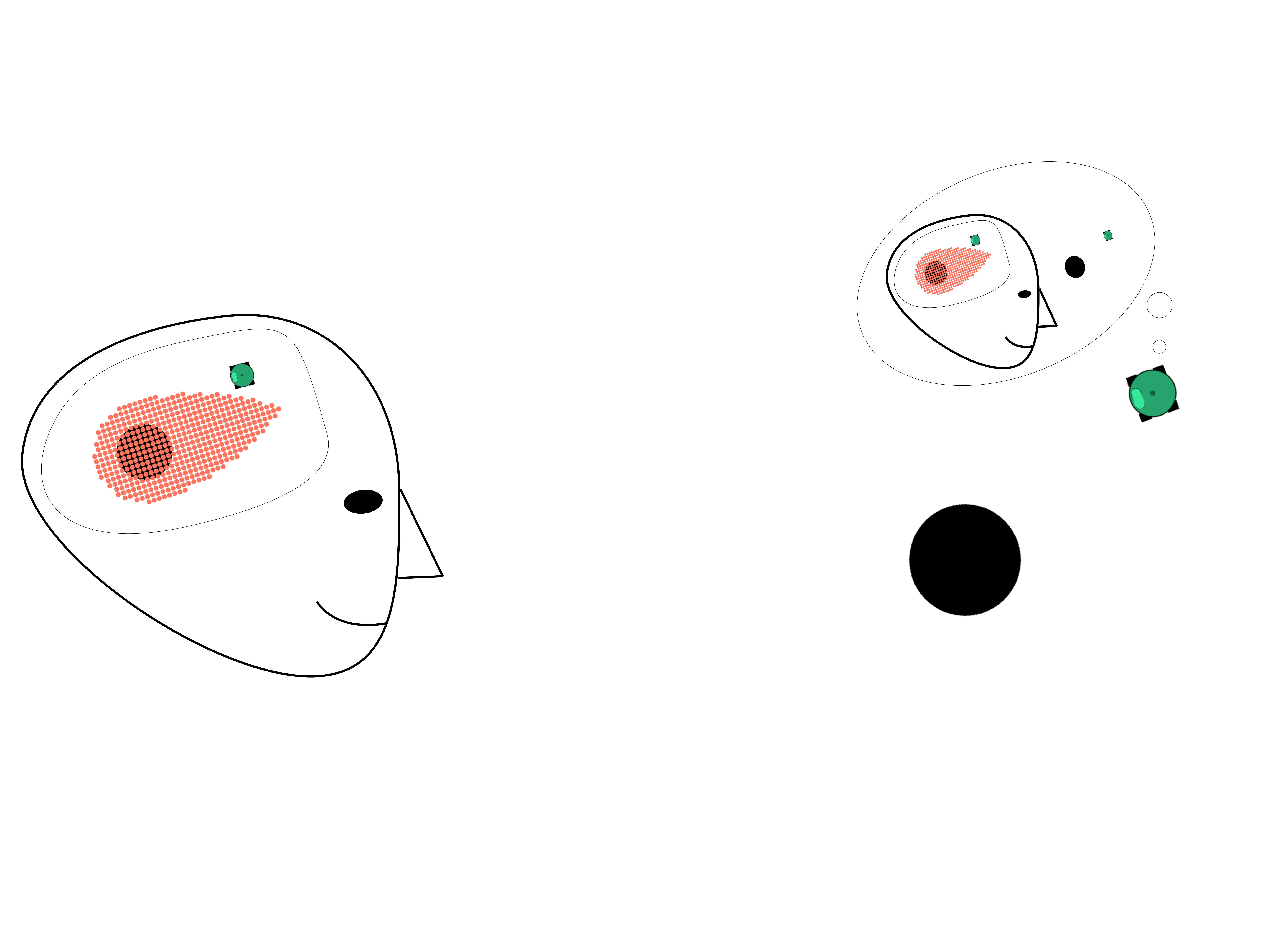}
  \caption{Top: if a robot's behavior does not take into account a human supervisor's notion of safety, the misaligned expectations can degrade team performance.
    Bottom: When a robot acts according to a human supervisor's expectations, the supervisor can more easily predict the robot's behavior.
  }
  \label{fig:intro_figure}
\end{figure}

While a human may be able to successfully exert direct control over a single robot, it becomes intractable for a human to directly control teams of robots (in fact, humans often benefit from automated assistance when controlling even a single robot, as discussed in the literature on assistive teleoperation \cite{dragan2013policy, javdani2017shared}).
In order to manage the increased complexity of multi-robot teams, the human must be able to rely on increased autonomy from the robots, freeing the human to focus their attention only on those areas where they are most needed.
Our goal is to model what grabs the supervisor's attention in order to modify robot behavior to reduce the occurrence of distractions.

This project is inspired by work like Bajcsy et al \cite{AndreaPush} and Jain et al \cite{coactiveLearning} that learn from supervisor interventions in a ``coactive'' learning framework.
These works apply Learning from Demonstration techniques to the more challenging domain where the given data is just a correction from a trajectory rather than a full trajectory.
The authors of \cite{AndreaPush} posed this correction challenge in model-based framework that interprets the human's signals as resulting from an optimization problem.
This inverse optimization framework has also been used in Inverse Reinforcement Learning \cite{abbeelIRL, ziebartIRL} which applies Inverse Optimal Control (as conceived of by Kalman \cite{kalmanInverseOptimalControl}) to interpreting human trajectories.
Our work applies the inverse optimization framework to learn from the supervisor's decisions to intervene.

Results in cognitive science suggest that humans observing physical scenes can be modeled as performing a noisy ``mental simulation'' to predict trajectories \cite{HamrickSim, StochasticMentalSim}.
We posit that human supervisors utilize this same cognitive dynamic simulation to predict robot safety and intervene accordingly.
Specifically, we theorize that the intervention behavior is driven by an internal ``safe set''  which we can attempt to reconstruct by observing supervisor interventions.

Safe sets are conceived from the Formal Methods notion of ``Viability''.
A set of states is ``viable'' if for every state in the set there exists a dynamic trajectory that stays within the set for all time.
Reachability analysis calculates the largest viable set that doesn't include any undesirable state configurations (e.g. collisions with obstacles, power overloads, etc).
Since the set is viable, it is possible to guarantee that the dynamic system will always stay within the set and therefore stay safely away from the undesirable states.
For this reason, viability kernels are often refered to as ``safe sets''.
Reachability can be used for robust path planning in dynamically changing environments \cite{fisac2015reach} or working around multiple dynamic agents \cite{chen2015safe}, and recent results have leveraged the technique to bound tracking error in order to generate dynamically feasible paths using simple planning algorithms \cite{FaSTrack}.

Hoffman et al. used the safety guarantees of reachability analysis to engineer a multi-drone team that could automatically avoid collisions \cite{hoffmannCollisionAvoidance}.
Similarly, Gillula could guarantee safety for learning algorithms by constraining their explorations to stay within the safe set \cite{gillulaMinimallyInvasive}.
Extending this, Akametalu and Tomlin \cite{KeneReachLearning} were able to guarantee safety while simultaneously learning and expanding the safe set.
All of these controllers supervise otherwise un-guaranteed systems and intervene to maintain safety whenever the system threatens to leave the viable safe set.
In this paper, we explore how this intervention behavior is similar to human supervision, and apply this to representing human safety concerns as safe sets in the state space.

\section{Supervisor Safe Set Control}
\label{sec:model_and_algorithm}
Based on the success of cognitive dynamical models for explaining humans' understanding of physical systems, we posit that human operators may have some notion of reachable sets which they employ to predict collisions or avoid obstacles.
We propose a noisy idealized model to describe the behavior of the human supervisor of a robotic team, and we develop a framework for estimating the human supervisor's mental model of a dynamical system based on observing their interactions with the team.
We then propose a control framework that capitalizes on this learned information to improve collaboration in such human-robot teams.

\subsection{Preliminaries: Reachability for Safety}
\label{sec:reachability_for_safety}
Consider a dynamical system with bounded input $u$ and bounded disturbance $d$, given by
\begin{align}
  \begin{gathered}
    \xdot = f(x, u, d), \\
    x \in \Rbb^n, \quad u \in \Ucal \subset \Rbb^{n_u}, \quad d \in \Dcal \subset \Rbb^{n_d},
  \end{gathered}
  \label{eq:bounded_system}
\end{align}
where $\Ucal$ and $\Dcal$ are compact.
We let $\Uscr$ and $\Dscr$ denote the  sets of measurable functions $\ubf : [0, \infty) \to \Ucal$ and $\dbf : [0, \infty) \to \Dcal$, respectively, which represent possible time histories for the system input and disturbance.
Given a choice of input and disturbance signals, there exists a unique continuous trajectory $\xi : [0, \infty) \to \Rbb^n$ from any initial state $x$ which solves
\begin{align}
  \begin{split}
    \dot{\xi}(t) &= f(\xi(t), \ubf(t), \dbf(t)), \, \text{a.e.} \ t \geq 0, \\
    \xi(0) &= x,
  \end{split}
  \label{eq:trajectory}
\end{align}
where $\xi(\cdot)$ describes the evolution of the dynamical system~\cite{coddingtonDiffEq}.

Obstacles in the environment can be modeled as a ``keep-out'' set of states $\Kcal \subset \Rbb^n$ that the system must avoid.
We define the safety of the system with respect to this set, such that the system is considered to be safe at state $\xi(0) = x$ over time horizon $T$ as long as we can choose $\ubf(\cdot)$ to guarantee that there exists no time $t \in [0, T]$ for which $\xi(t) \in \Kcal$.
The task of maintaining the system's safety over this interval can be modeled as a differential game between the control input and the disturbance.
Consider an optimal control signal $\ubf(\cdot)$ which attempts to steer the system away from $\Kcal$ and an optimal disturbance $\dbf(\cdot)$ which attempts to drive the system towards $\Kcal$.
By choosing any Lipschitz payoff function ${l : \Rbb^n \to \Rbb}$ which is negative-valued for $x \in \Kcal$ and positive for $x \notin \Kcal$, we can encode the outcome of this game via a value function $V(x, t)$ characterized by the following Hamilton-Jacobi-Isaacs variational inequality~\cite{fisacSafetyFramework}:
\begin{align}
  \begin{split}
    &\min \begin{cases} l(x) - V(x, t), \\ \pd{V}{t}(x, t) + \max\limits_{u \in \Ucal} \min\limits_{d \in \Dcal} \pd{V}{x}(x, t) \cdotp f(x,u,d)
      \end{cases} = \ 0 \\
    &V(x, T) = l(x).
  \end{split}
  \label{eq:diff_game}
\end{align}

The value function $V(x, t)$ that satisfies the above conditions is equal to $\min_{\tau \in [t, T]}l(\xi^*(\tau))$ for the trajectory with $\xi^*(t) = x$ driven by an optimal control $\ubf(\cdot)$ and an optimal disturbance $\dbf(\cdot)$.
We can therefore find the set of states ${\Rcal_T = \left\{ x \in \Rbb^n \suchthat V(x, 0) < 0 \right\}}$ from which we cannot guarantee the safety of the system on the interval $[0, T]$, also known as the backward-reachable set of $\Kcal$ over this interval.
That is, for all initial states $x \in \Rcal_T$ and feedback control polices $\ubf(t) = g(\xi(t))$, there exists some disturbance $\dbf(\cdot) \in \Dscr$ such that $\xi(t) \in \Kcal$ for some $t \in [0, T]$.

\begin{figure}[t]
  \vspace{10pt}
  \centering
  \includegraphics[width=0.7\columnwidth]{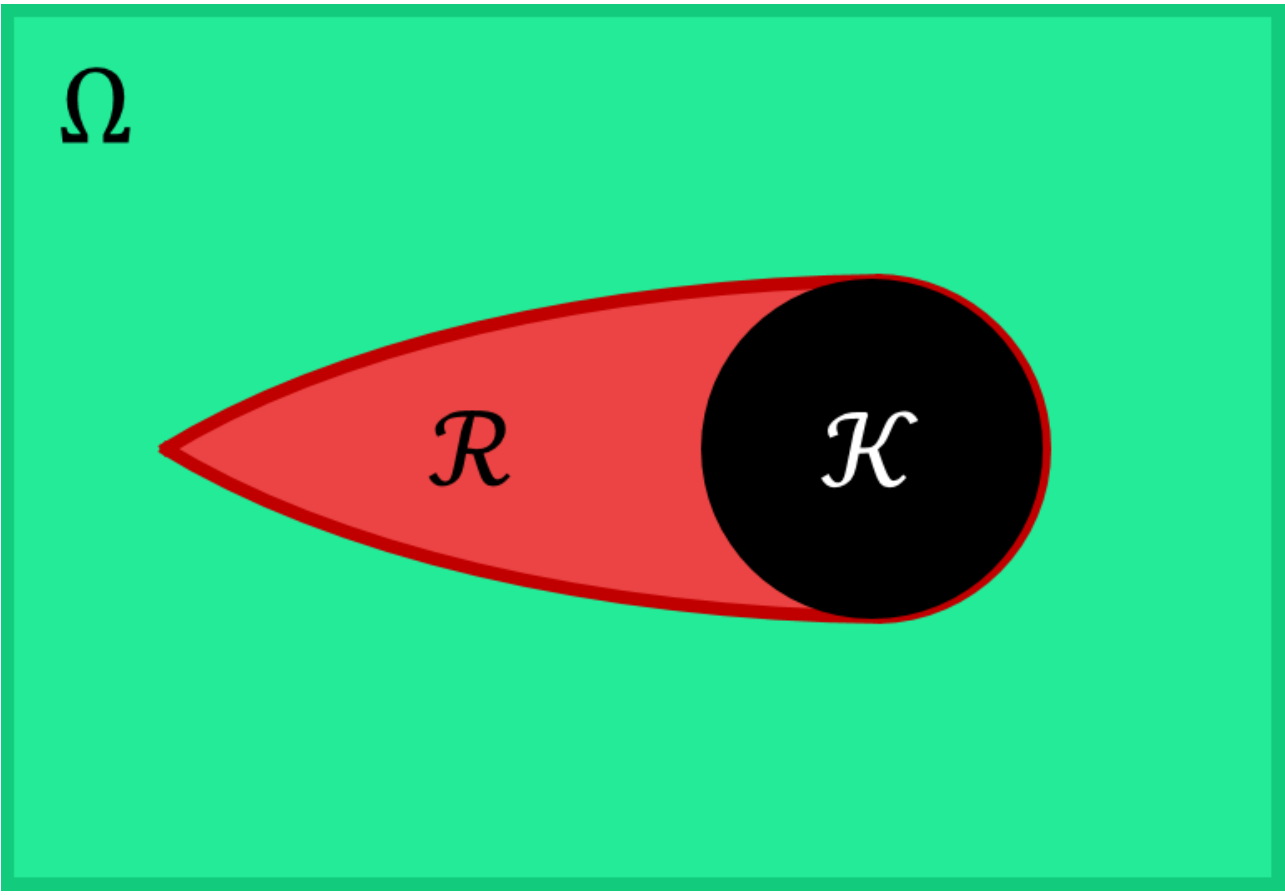}
  \caption{Illustration of the relationship between a keep-out set $\Kcal$, the derived backward-reachable set $\Rcal$, and the resulting safe set $\Omega$.
    Note that $\Kcal \subseteq \Rcal$, and $\Omega$ is equal to the complement of $\Rcal$.
    This illustration approximates the result obtained using the Dubins car dynamics given in \eqref{eq:dubins}.}
  \label{fig:krw}
  \vspace{-10pt}
\end{figure}

If there exists a non-empty controlled-invariant set $\Omega$ that does not intersect $\Kcal$, then we deem this set $\Omega$ a ``safe set'' because there exists a feedback policy that guarantees that the system remains in $\Omega$, and thus out of $\Kcal$, for all time.
It follows from their properties that $\Omega$ is the complement of $\Rcal_T$, and the relationship between $\Kcal$, $\Rcal_T$, and $\Omega$ is visualized in Fig. \ref{fig:krw}.
Within a safe set $\Omega$, the value function becomes independent of $t$ as $T \to \infty$ \cite{fisacSafetyFramework}.
Because we focus on the case where the system is initialized to some safe state $\xi(0) \in \Omega$ and we aim to maintain $\xi(t) \in \Omega$ for all $t \in [0, \infty)$, we simplify notation by defining the terms $V(x) \triangleq \lim_{T \to \infty} V(x, \cdot)$ and $\Rcal \triangleq \Rcal_\infty$.

One approach to guaranteeing the safety of the system is to apply a ``minimally invasive'' controller which activates on the zero level set of $V(x)$ \cite{gillulaMinimallyInvasive}.
This approach allows complete flexibility of control as long as $\xi(t) \in \text{interior}(\Omega)$, and applies the optimal control to avoid $\Kcal$ when $\xi(\cdot)$ reaches the boundary of $\Omega$.
We refer the interested reader to \cite{gillulaMinimallyInvasive, fisacSafetyFramework} for a more thorough treatment of reachability and minimally invasive controllers.


\subsection{Noisy Idealized Supervisor Model}
\label{sec:idealized_supervisor}
We define an idealized model of the supervisor of a robotic team whose responsibility it is to ensure that no robots collide with obstacles represented by the keep-out set $\Kcal$.
The idealized supervisor behaves as a minimally invasive controller as described in Section \ref{sec:reachability_for_safety}.
However, while the robotic team members' true dynamics are given by the function $f(x, u, d)$ as in \eqref{eq:bounded_system}, the supervisor possesses an internal model of the robots' dynamics given by $f_S(x, u, d)$, which is not necessarily equal to the true dynamics.
Following the differential game characterized by \eqref{eq:diff_game}, the supervisor also possesses an internal value function $V_S(\cdot)$ and safe set $\Omega_S$ which they use to evaluate the safety of each state $x$ in the environment.
We allow for the possibility that the supervisor adds some amount of margin $\mu$ to their internal safe set, such that $\Omega_S = \left\{ x \in \Rbb^n \suchthat V_S(x) \geq \mu \right\}$.
Therefore, the idealized supervisor will always intervene when a robotic team member reaches the $\mu$ level set of $V_S(\cdot)$, rather than the zero level set of the true $V(\cdot)$.
We further specify that the idealized supervisor is \emph{conservative}: ${\forall x \in \Rbb^n, \ V(x) \leq 0 \implies V_S(x) \leq \mu}$. This condition implies that the supervisor will never let a robot teammate leave the true safe set $\Omega$ since $\Omega_S \subseteq \Omega$.
Additionally, we propose a noisy version of this idealized supervisor: the noisy idealized supervisor will intervene when they observe a robot reach the $\mu + w$ level set of $V_S(\cdot)$, where $w$ is drawn from $\Ncal(0,\sigma_S^2)$ whenever a supervisor makes a safety judgement.


\subsection{Learning Safe Sets from Supervisor Interventions}
\label{sec:learn_safe_sets}
We choose to model the human supervisor of a robotic team as approximating the behavior of the idealized supervisor model presented in Section \ref{sec:idealized_supervisor}.
That is, the human supervisor will allow the robots to perform their task however they choose, but intervene whenever they \emph{perceive} that a robot is approaching an obstacle $\Kcal$ in the state space.
Given this model, we can interpret the points at which the human intervenes as corresponding to the unknown $\mu$ level set of some value function $V_H(\cdot): \Rbb^n \to \Rbb$, which characterizes the human's mental safe set $\Omega_H$.
Our goal is to use observations of human interventions to derive an estimated value function $\Vhat_H(\cdot)$ and $\hat{\mu}$ which describe the observed behavior and induce an estimated $\hat{\Omega}_H$.
We approach this task by deriving a Maximum Likelihood Estimator (MLE) of the human's mental safe set.
If we assume that a human supervisor always intends to intervene at the $\mu$ level set of $V_H(x)$, but their ability to precisely intervene at this level is subject to Gaussian noise, either from observation error or variability in reaction time, then we can consider the value at an intervention point $x_i$ as being drawn from a normal distribution centered at $\mu$ (that is, $V_H(x_i) \sim \Ncal(\mu, \sigma^2$)).

Given a proposed value function $\Vhat_H(\cdot)$ and a set of intervention points $\{x_1, x_2, \cdots, x_p\}$ with corresponding values $\{\Vhat_H(x_1), \Vhat_H(x_2), \cdots, \Vhat_H(x_p)\}$, we wish to estimate the most likely $\mu$ and $\sigma^2$ to explain these interventions.
Gaussian distributions induce the following probability density function for a single observation $\Vhat_H(x_j)$
\begin{align}
  f\left( \Vhat_H(x_j) \ |\ \mu, \sigma^2 \right) = \frac{1}{\sqrt{2\pi\sigma^2}}\exp\left(-\frac{\left(\Vhat_H(x_j) - \mu \right)^2}{2\sigma^2}\right)
  \label{eq:gaussian_pdf}
\end{align}
which leads to the following probability density for a set of $p$ independent observations
\begin{align}
  \begin{split}
    &f\left(\Vhat_H(x_1), \cdots, \Vhat_H(x_p) \ | \ \mu, \sigma^2 \right) = \prod_{j = 1}^{p} f\left( \Vhat_H(x_j) \ |\ \mu, \sigma^2 \right) \\
    &= \left(\frac{1}{2\pi\sigma^2}\right)^{\frac{p}{2}}
      \exp\left(-\frac{\sum_{j= 1}^{p} \left(\Vhat_H(x_j) - \mu \right) ^2}{2\sigma^2}\right).
  \end{split}
  \label{eq:gaussian_pdf_multi}
\end{align}
The likelihood of any estimated parameter values $\hat{\mu}$ and $\hat{\sigma}^2$ being correct, given the observations and the proposed value function $\Vhat_H(\cdot)$, is expressed as ${\Lcal\left(\hat{\mu}, \hat{\sigma}^2 \ | \ \Vhat_H(\cdot)\right) = f\left(\Vhat_H(x_1), \cdots, \Vhat_H(x_p) \ | \ \hat{\mu}, \hat{\sigma}^2 \right)}$.
It can be shown that the values of the unknown parameters $\mu$ and $\sigma^2$ that maximize the likelihood function are given by
\begin{align}
  \hat{\mu}^* = \frac{1}{p}\sum_{j = 1}^{p} \Vhat_H(x_j) &\quad\text{and}& \hat{\sigma}^{*2} = \frac{1}{p}\sum_{j = 1}^{p} \left(\Vhat_H(x_j) - \hat{\mu}^* \right)^2,
  \label{eq:sample_parameters}
\end{align}
which are simply the mean and variance of the set of observations.

Notice that the estimates given by \eqref{eq:sample_parameters} are computed with respect to a given value function $\Vhat_H(\cdot)$.
If we were to assume that the human supervisor has a perfect model of the system dynamics, then we could simply set $\Vhat_H(\cdot)$ to equal the true $V(\cdot)$ of the system in \eqref{eq:bounded_system}, and $\hat{\mu}^*$ would be the maximum likelihood estimate for the level at which the supervisor will intervene.
However, it is unlikely that a human supervisor's notion of the dynamics will correspond exactly to this model, and we would like to maintain the flexibility of estimating value functions that are not strictly derived from \eqref{eq:bounded_system}.
To this end, we define the maximum likelihood of $\Vhat_H(\cdot)$ being the $V_H(\cdot)$ that produced our observations as $\Lcals(\Vhat_H(\cdot)) = \max_{\hat{\mu}, \hat{\sigma}^2}\Lcal(\hat{\mu}, \hat{\sigma}^2 \ | \ \Vhat_H(\cdot))$.
The value of $\Lcals(\Vhat_H(\cdot))$ is obtained by substituting the estimates from \eqref{eq:sample_parameters} into the probability density function from \eqref{eq:gaussian_pdf_multi}.
That is, $\Lcals \left(\Vhat_H(\cdot)\right) = f\left(\Vhat_H(x_1), \cdots, \Vhat_H(x_p) \ | \ \hat{\mu}^*, \hat{\sigma}^{*2}\right)$.

We seek the most likely value function to explain our observations, which will be the value function $\Vhat^*(\cdot)$ with the greatest maximum likelihood $\Lcals(\Vhat^*(\cdot))$ (the maximum over maxima)
\begin{align}
  \Vhat^*(\cdot) = \argmax_{V(\cdot) \in \Vcal} \Lcals \left( V(\cdot) \right),
  \label{eq:max_over_values}
\end{align}
where $\Vcal$ is the set of all possible value functions.

In order to make this optimization tractable, we can restrict ourselves to a set of value functions $\{V_\theta(\cdot)\}_{\theta \in \Rbb^m}$ corresponding to a family of dynamics functions $\{f_\theta(\cdot, \cdot, \cdot) \}_{\theta \in \Rbb^m}$ parameterized by $\theta \in \Rbb^m$, making the optimization in question
\begin{align}
  \Vhat^*(\cdot) = \argmax_{\theta \in \Rbb^m} \Lcals \left( V_\theta(\cdot) \right).
  \label{eq:max_over_theta}
\end{align}

\begin{figure}[t]
  \centering
  \includegraphics[width=\columnwidth]{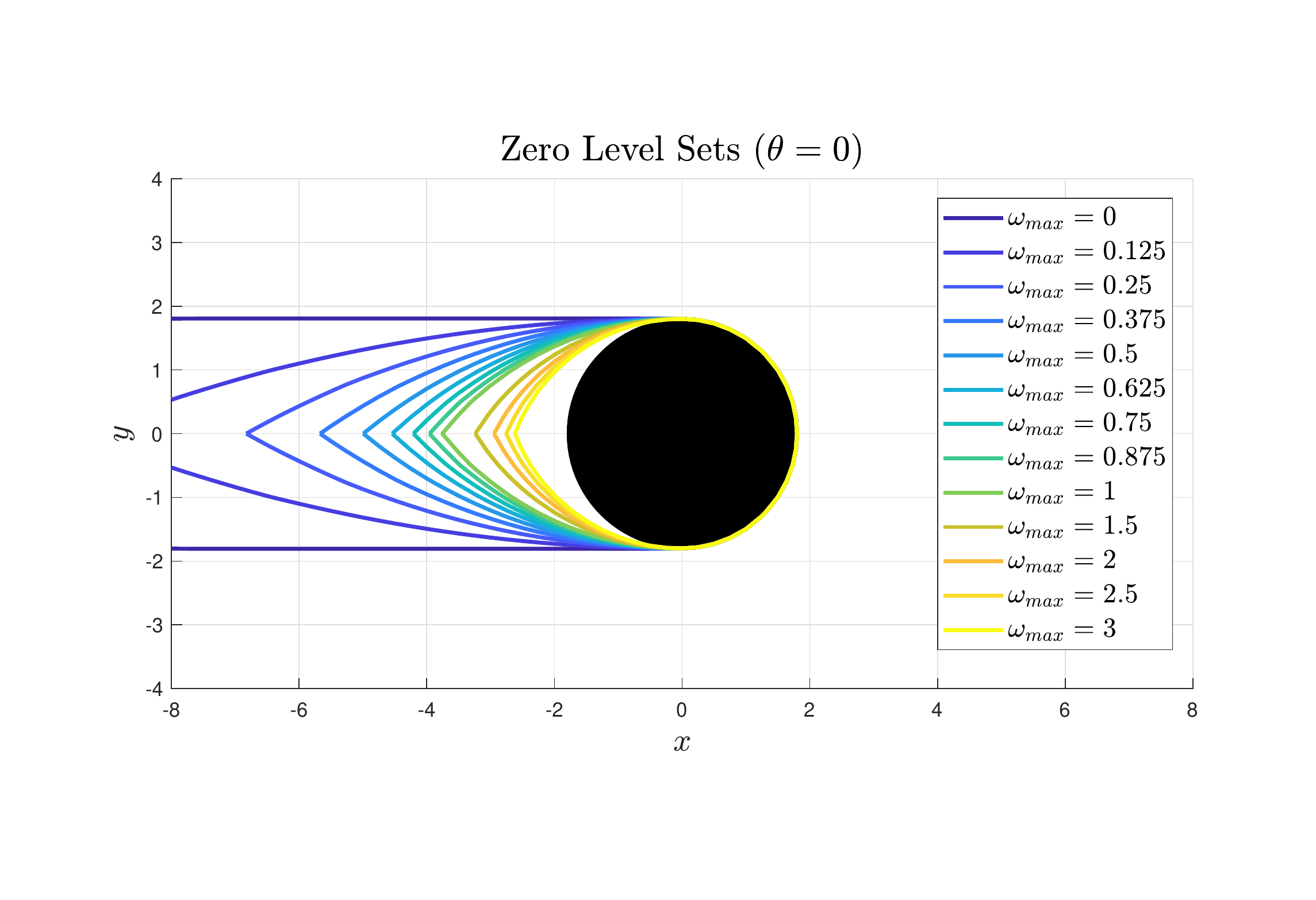}
  \caption{Two dimensional slices of the zero level sets of the value functions $V_i(\cdot)$ from the library used for the experiment described in Section \ref{sec:experimentalDesign}. We used a family of Dubins car dynamics (see \eqref{eq:dubins}) parametrized by $\omega_{max}$. Notice that as $\omega_{max}$ decreases (the modeled control authority is decreased), the level sets extend farther away from the obstacle, indicating that a robot is expected to turn earlier to guarantee safety.}
  \label{fig:zero_level_sets}
\end{figure}

In practice, we may not be able to find an expression for the gradient of $\Lcals(V_\theta(\cdot))$ with respect to $\theta$, since the value function is derived from the dynamics $f_\theta(\cdot, \cdot, \cdot)$ via the differential game given by \eqref{eq:diff_game}.
The lack of a gradient expression restricts the use of numerical methods to solve the problem as presented in \eqref{eq:max_over_theta}.
In these cases, we can compute a representative library of $b$ value functions $\{V_i(\cdot)\}_{i = 1}^{b}$ corresponding to a set of $b$ representative parameter values $\{\theta_i\}_{i = 1}^{b}$ (see Fig. \ref{fig:zero_level_sets} for an example library).
The optimization then reduces to choosing the most likely value function from among this library
\begin{align}
  \Vhat^*(\cdot) = \argmax_{i \in \{1, \cdots, b\}} \Lcals \left( V_i(\cdot) \right).
  \label{eq:max_over_i}
\end{align}

\begin{figure}[t]
  \vspace{10pt}
  \centering
  \includegraphics[width=\columnwidth]{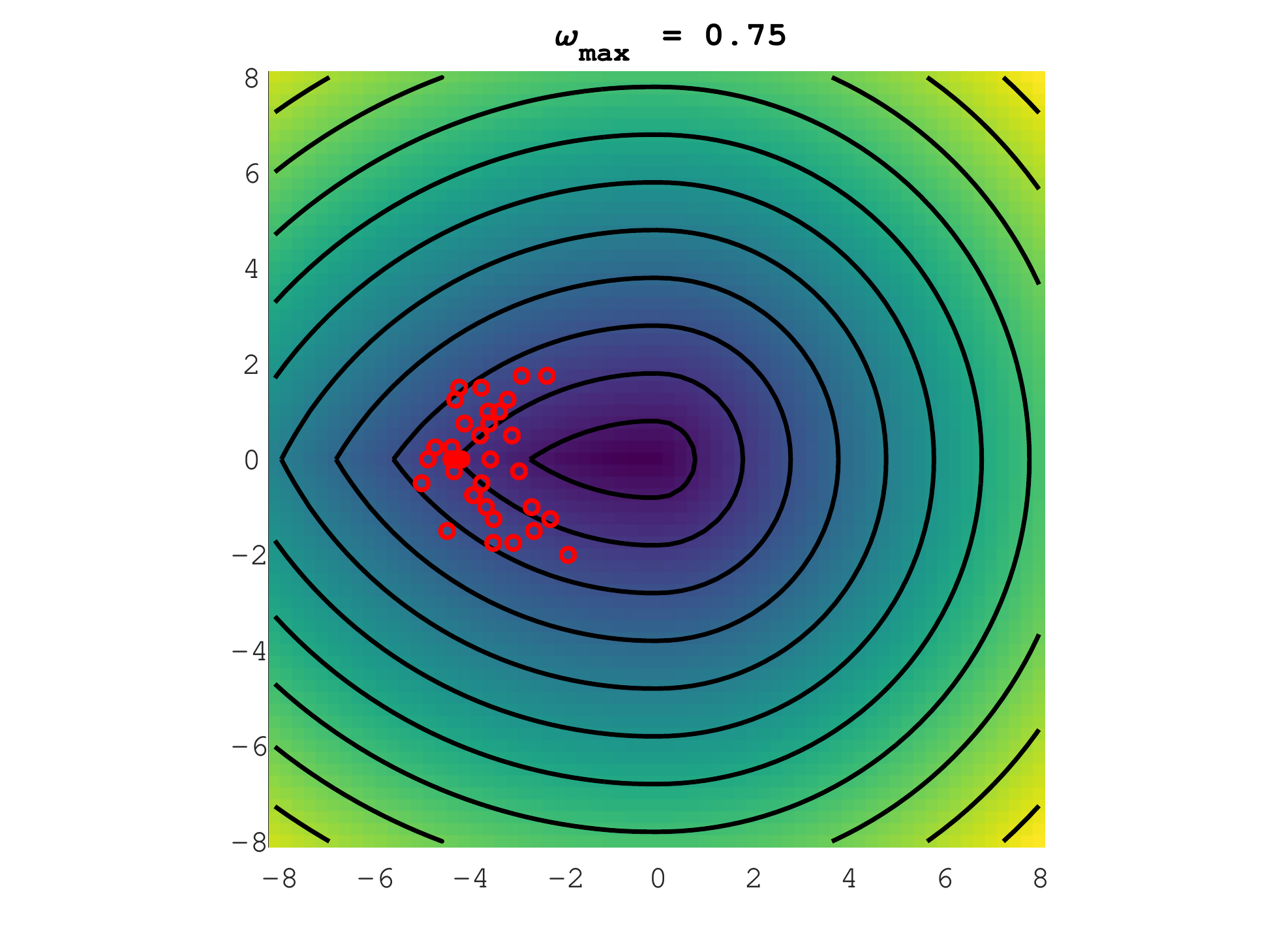}
  \caption{An example data set from the experiment described in Section \ref{sec:experimentalDesign}. The red circles represent the location of supervisor interventions, and the colored background represents the learned value function $V(\cdot)$ with contour lines shown in black. In this case, the learning algorithm chose a dynamics model parametrized by $\omega_{max} = 0.75$.}
  \label{fig:flinch_learning}
  \vspace{-10pt}
\end{figure}

In order to ensure that the learned safe set is conservative, we can extend our MLE to a Maximum A Posteriori (MAP) estimator by incorporating our prior belief that, regardless of the safe set that the supervisor uses to generate interventions, they do not want the robots to be unsafe with respect to the true dynamics.
In this case, we maintain a uniform prior $P(\theta)$ that assigns equal probability to all $V_\theta(\cdot)$ whose zero sublevel sets are supersets of the zero sublevel set of the true $V(\cdot)$, and zero probability to all other $V_\theta(\cdot)$.
In other words, we assume that the supervisor does not overestimate the agility of the robots, and in practice we can enforce this condition by choosing the library in \eqref{eq:max_over_i} to only contain appropriate value functions.
Moreover, regardless of the choice of $\Vhat_H(\cdot)$, we assume that the supervisor intends to intervene before reaching the zero level set of $\Vhat_H(\cdot)$, which always includes the boundary of $\Kcal$.
If we choose a prior $P(\mu)$ that assigns zero probability to all non-positive $\mu$ and uniform probability elsewhere, it can be shown that the MAP estimates are obtained by letting $\hat{\mu}^*$ equal $\max\left\{ \hat{\mu}^*,\, 0 \right\}$ and otherwise proceeding as before.
Fig. \ref{fig:flinch_learning} provides an example of this algorithm estimating a safe set from human supervisor intervention data.

\subsection{Team Control with Learned Safe Sets}
\label{sec:team_control}
We propose that safe sets learned according to the approach in Section \ref{sec:learn_safe_sets} can be used to create effective control laws for the robotic members of human-robot teams.
Recall our model of the human supervisor of a robotic team: the supervisor must rely on each robot's autonomy to complete the majority of their tasks unassisted, but the supervisor may intervene to correct a robot's behavior when necessary (such as by avoiding an imminent collision with the keep-out set $\Kcal$).
We put forth that in the scenario where the human intervenes to prevent a collision, they do so because they observe that a robot has violated the boundaries of their mental safe set $\Omega_H$.

Now, consider a team of robots navigating an unknown environment, and which are able to avoid any obstacles that they detect.
One approach to safely automating this team is to have each robot behave according to a minimally invasive control law: the robots are allowed to follow trajectories generated by any planning algorithm, so long as they remain within $\Omega$, the reachable set computed using the baseline dynamics model \eqref{eq:bounded_system} with associated value function $V(\cdot)$.
Whenever these robots detect an obstacle, they add it to the keep-out set $\Kcal$, thus modifying $\Omega$ and $V(\cdot)$.
If a robot reaches the boundary of $\Omega$, it applies the optimal control to avoid $\Kcal$ until it has cleared the obstacle.
However, it is possible that a robot does not detect an obstacle, and a human supervisor must intervene to ensure robot safety.

As stated above, the human supervisor will intervene when a robot reaches the boundary of $\Omega_H$, not the boundary of $\Omega$.
This discrepancy leads to the possibility that the supervisor will intervene when the robot reaches some state $x$, even if the robot would have avoided the obstacle without intervention.
These situations arise whenever $V_H(x) \leq \mu$ but $V(x) > 0$.
These ``false positive'' interventions represent unnecessary work for the human supervisor, and we seek to eliminate them in order to improve the human's experience and the team's overall performance.

We propose using a safe set $\hat{\Omega}_H$ learned from previous observations of supervisor interventions, as outlined in Section \ref{sec:learn_safe_sets}, as a substitute for $\Omega$ in the robots' minimally invasive control law.
By estimating the human's internal safe set, we take advantage of the following property:
\begin{property}
  For an idealized supervisor collaborating with a team of robots as described in Section \ref{sec:team_control}, if the robots avoid detected obstacles $\Kcal$ by applying an optimally safe control at the boundary of safe set $\Omega_S$, then if the supervisor plans to intervene because they observe $\xi_i(t) \in R_S$ for robot $i$, the supervisor can infer that robot $i$ has not detected an obstacle and any supervisor intervention will not be a false positive.
\end{property}
\begin{proof}
  The proof of this property follows constructively from the definitions of safe set, idealized supervisor, and false positive.
  If robot $i$ had correctly detected an obstacle and adjusted its representation of $\Omega_S$ accordingly, then it would have applied the optimal control to remain within the supervisor's safe set.
  Therefore, if the supervisor is able to observe that robot $i$ has left $\Omega_S$, it must be the case that the robot has not detected the obstacle.
  False positives are defined to be supervisor interventions that occur when a robot has detected an obstacle but the supervisor still intervenes.
  In this case, the supervisor can correctly infer that robot $i$ has not detected an obstacle, so any intervention at this point cannot be a false positive.
\end{proof}

For an idealized supervisor, as $\hat{\Omega}_S$ becomes an arbitrarily good approximation of $\Omega_S$, the number of false positive interventions will approach zero.
For a \emph{noisy} idealized supervisor, the supervisor will intervene whenever $V_S(x)+w\leq\mu$ where $w\sim\Ncal(0,\sigma_S^2)$.
This noise will continue to produce false positives, even with a perfect fit $\hat{\Omega}_S  = \Omega_S$, if the robots apply the optimally safe control at the $\mu$-level set of $\Omega_S$.
Instead, the level set $\alpha$ where the optimally safe control is applied can be raised arbitrarily high to drive the false positive rate to zero.
For example, $\alpha = \mu + 2 \sigma_S$ is sufficient to avoid over 97\% of intervention states used for learning, in expectation.
We test the efficacy of our approach through the human-subjects experiment described in Section \ref{sec:experimentalDesign}.



\section{Experimental Design for User Validation}
\label{sec:experimentalDesign}
Our goal in understanding and modeling the supervisor's conception of safety is to improve team performance by decreasing cognitive overload.
Although we have based our human modeling on the cognitive science literature, we do not intend to verify humans' exact cognitive processes.
Instead, we aim to apply our inspiration from cognitive science toward building better human-robot teams.
To this end, our hypotheses are:
\subsubsection{H1} Representing supervisor behavior as cognitive keep-out sets allows intervention signals to be distilled into an actionable rule which will decrease supervisory false positives and cognitive strain, thereby increasing team performance and trust.
\subsubsection{H2} Fitting danger-avoidance behavior to a supervisor's beliefs is preferable to generic conservative behavior.

In our experiment, we gather supervisor intervention data, fit our model to the data, and then run a human-robot teaming task that assesses performance.

\begin{figure}[t]
  \vspace{10pt}
  \centering
  \includegraphics[width=0.98\columnwidth]{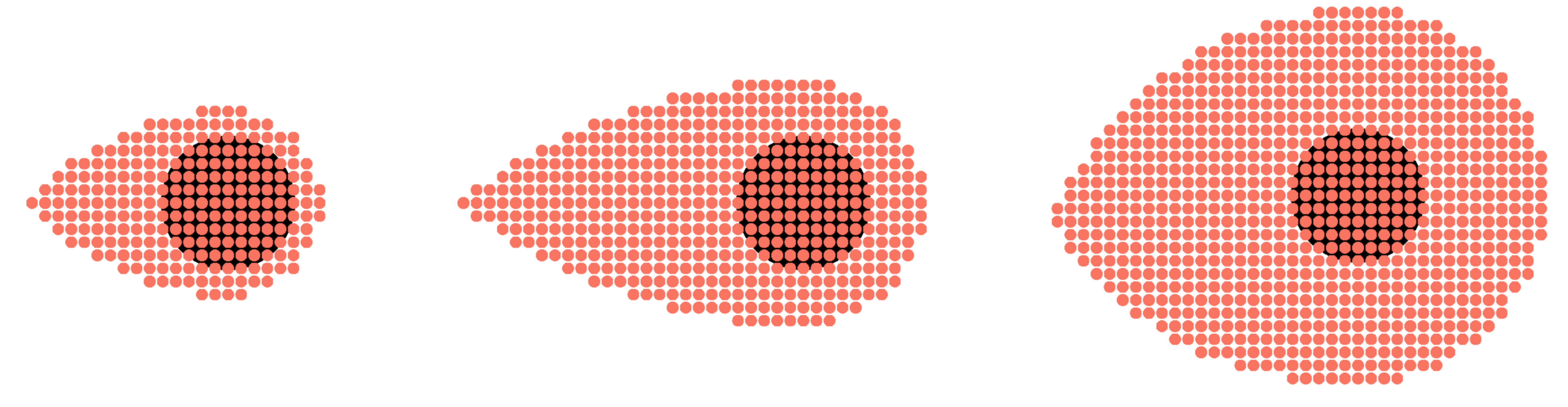}
  \caption{Safe sets tested in experiment (illustrated by their complementary reachable set):
    (left) Standard safe set (calculated from true dynamics and obstacle size),
    (middle) example Learned safe set (calculated from fitted supervisory perception of dynamics and obstacle size),
    (right) Conservative safe set (calculated from true dynamics and inflated obstacle size)
  }
  \label{fig:tested_safesets}
  \vspace{-10pt}
\end{figure}

\subsection{Procedure}
Our experiment applies the idealized supervisor theory and learning algorithm to supervising simulated robots.
The robots moved according to the Dubins car model:

\begin{equation}
  \begin{gathered}
    \begin{split}
      \dot{x} &= 3 \cos(\theta) \\
      \dot{y} &= 3 \sin(\theta) \\
      \dot{\theta} &= u
    \end{split} \\
    u \in \Ucal = [-\omega_{max}, \omega_{max}], \ \omega_{max} = 1
  \end{gathered}
  \label{eq:dubins}
\end{equation}

The experiment is divided into three phases.
In Phase I, the subject is given an opportunity to familiarize themselves with the robotic system's dynamics.
The user is allowed to directly apply the full range of controls through the computer keyboard for one minute.
After ensuring the user has some experience from which to build an internal dynamics model, we then assess their emergent conception of safety.
In Phase~II, supervisory data is extracted from the subject by showing them scenes where the robot is driving towards an obstacle, and the supervisor decides where to intervene to avoid a crash.
This intervention data is then fed into our algorithm (described in Section \ref{sec:learn_safe_sets}) that extracts the best fitting safe set.
Our estimator used a library of candidate dynamics functions parameterized by values of $\omega_{max}$ between 0 and 3, as shown in Fig. \ref{fig:zero_level_sets}.
In this experiment, we enforced conservativeness by excluding subjects whose Learned sets were not supersets of the Standard safe set, rather than enforcing a prior directly on $\omega_{max}$.
The Learned safe set is assessed in Phase III against two fixed safe sets (see Fig. \ref{fig:tested_safesets}) pre-calculated from the true dynamic equations.

\begin{figure}[t]
  \vspace{10pt}
  \centering
  \includegraphics[width=\columnwidth]{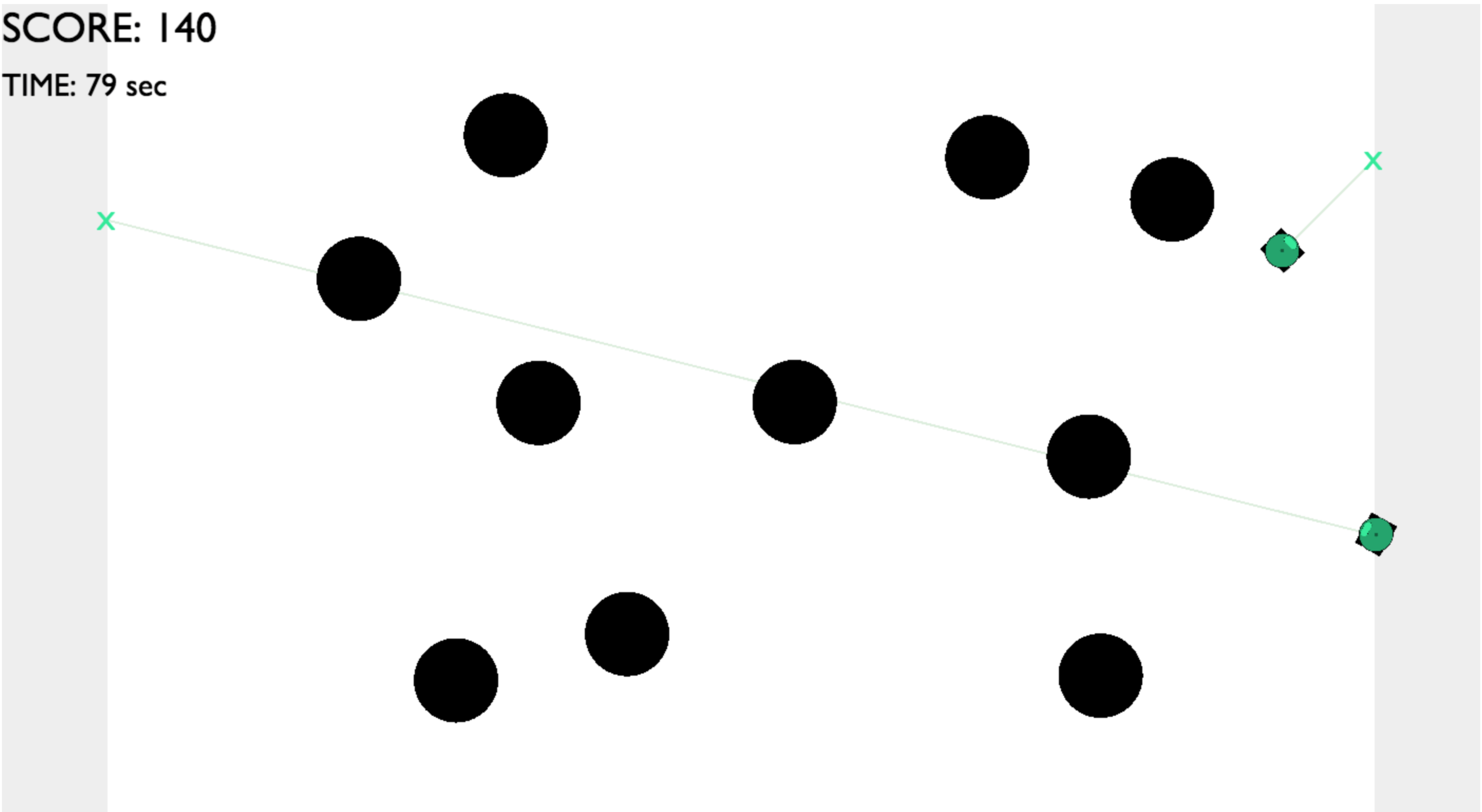}
  \caption{
    Screenshot of the task from Phase III of the experiment.
    Robotic vehicles make trips back and forth across the screen, detecting and avoiding each obstacle with 80\% probability.
    The human supervisor must remove an obstacle in the event that it is undetected, but must infer this information from the robots' motion.
  }
  \label{fig:ExperimentScreenshot}
  \vspace{-10pt}
\end{figure}

These safe sets were calculated using Hamilton-Jacobi reachability as described in Section \ref{sec:reachability_for_safety} using the Level Set Toolbox \cite{mitchell2004toolbox} for MATLAB.
During this final phase, the subject sequentially supervises homogeneous teams of robots, each team avoiding obstacles based on one of the three assessed safe sets.
Ten randomly placed obstacles are strewn about the screen impeding the robots' autonomous trips back and forth across the screen (see Fig. \ref{fig:ExperimentScreenshot}).
Although robots will detect and avoid an obstacles in 80\% of their interactions with it, there is a 20\% chance that the robot will not detect an obstacle as it approaches.
The subject is charged with catching these random failures and removing an obstacle before the robot crashes.
Crashing is disincentivized by decrementing an on-screen ``score'' counter.
Removing an obstacle costs only half of what a crash costs the player.
This system encourages saving the robot but not guessing wildly.
Moreover, simply clearing out all obstacles is not a viable strategy because every obstacle removed generates a new obstacle elsewhere.
This score mechanism was also used to make the participant invested in team success by awarding points every time a robot completes a trip across the screen.

\subsection{Independent Variables}
To assess our hypotheses, we manipulate the safe set used between team supervision trials.
We exposed the human subject to three teams, each driving using one of three safe sets.
The Learned set is derived from Phase II supervisor intervention observations as described in Section \ref{sec:model_and_algorithm}, using ${\alpha = \mu}$.
The two baseline kernels are calculated using Hamilton-Jacobi-Isaacs reachability on the true dynamic equations.
The Standard set is calculated using the true obstacle size.
The Conservative set adds a buffer that doubles the effective size of the obstacle, inducing trajectories that give obstacles a wide berth.

\subsection{Dependent Measures}
\subsubsection{Objective Measures}
The team was tasked with making trips across the screen to reach randomized goals.
The robots' task was to travel across the screen, safely dodging obstacles along the way, while the human was tasked with supervising as a failsafe to remove an obstacle if the robots should fail to observe and avoid it.

Team performance was quantified using three objective metrics: number of trips completed, number of supervisory interventions, and the number of obstacle collisions.
These metrics were presented to the subject as an aggregated, arcade-style score.
To incentivize participants to only intervene when necessary, obstacle-removal interventions reduced the score, but only by half as much as an obstacle collision.

The number of interventions taken by the supervisor can also serve as a proxy measurement to quantify the amount of cognitive strain they experience while working with the robotic team.
Of particular note are the number of interventions that were not actually required, as the supervisor incorrectly judged that a robot had not detected an obstacle.
These false positives needlessly drain supervisor attention and indicate a lack of trust in the system.
We aim to increase the human's trust in the system, which we quantify by a decrease in these false positives.

\subsubsection{Subjective Measures}
After each round of pairwise comparison (completing the task with two different robotic teams), we presented the subject with a questionnaire to gauge how the choice of safe set impacted their experience.
These questionnaires contained statements about each team that subjects would respond to using a 7-point Likert scale (1 - Strongly Disagree, 7 - Strongly Agree).
These statements were designed to measure Trust, Perceived Performance, Interpretability, Confidence, Team Fluency, and overall Preference between the teams in the comparison.

\subsection{Subject Allocation}
The subject population consisted of 6 male, 5 female, and 1 non-binary participants between the ages of 18-29.
We used a within-subjects design where each subject was asked to complete all three possible pairwise comparisons of our three treatments (the safe sets used).
We used a balanced Latin Square design for the order of comparisons, with no treatment being first in a pair twice.
Furthermore, we generated six randomized versions of the task so that subjects were presented with a different version of the task for each trial across the three pairwise comparisons.
To avoid coupling the treatment results to a particular version of the task, each treatment was paired with each task version an equal number of times across our subject population.


\section{Analysis and Discussion}
\label{analysis_and_discussion}
\subsection{H1: False Positive Reduction over Standard}
\label{sec:analysisH1}
Our first hypothesis is that a Learned safe set that reflects the supervisor's intervention behavior would decrease the number of false positives compared to the Standard safe set.
To test this, we performed a one-way repeated measures ANOVA on the number of supervisory false positives from Phase III of the experiment with safe set as the manipulated factor.
A false positive was any supervisor intervention where the removed obstacle was actually detected by all nearby robots, which would have avoided it successfully.
The robot team's safe set had a significant effect on the number of supervisory false positives ($F(2,20) = 8.72$, $p < 0.01$).
An all-pairs post-hoc Tukey method found that the Learned safe set significantly decreased ($p = 0.0328 < 0.05$) false positives over the Standard safe set, but there was no significant difference between the Learned safe set and the Conservative safe set (which also significantly decreased false positives over the Standard safe set, with $p < 0.01$).
These results support our main hypothesis that \textbf{\textit{representing supervisor behavior as cognitive keep-out sets allows intervention signals to be distilled into an actionable rule which will decrease supervisory false positives}}.

\begin{figure}[t]
  \vspace{10pt}
  \centering
  \includegraphics[width=\columnwidth]{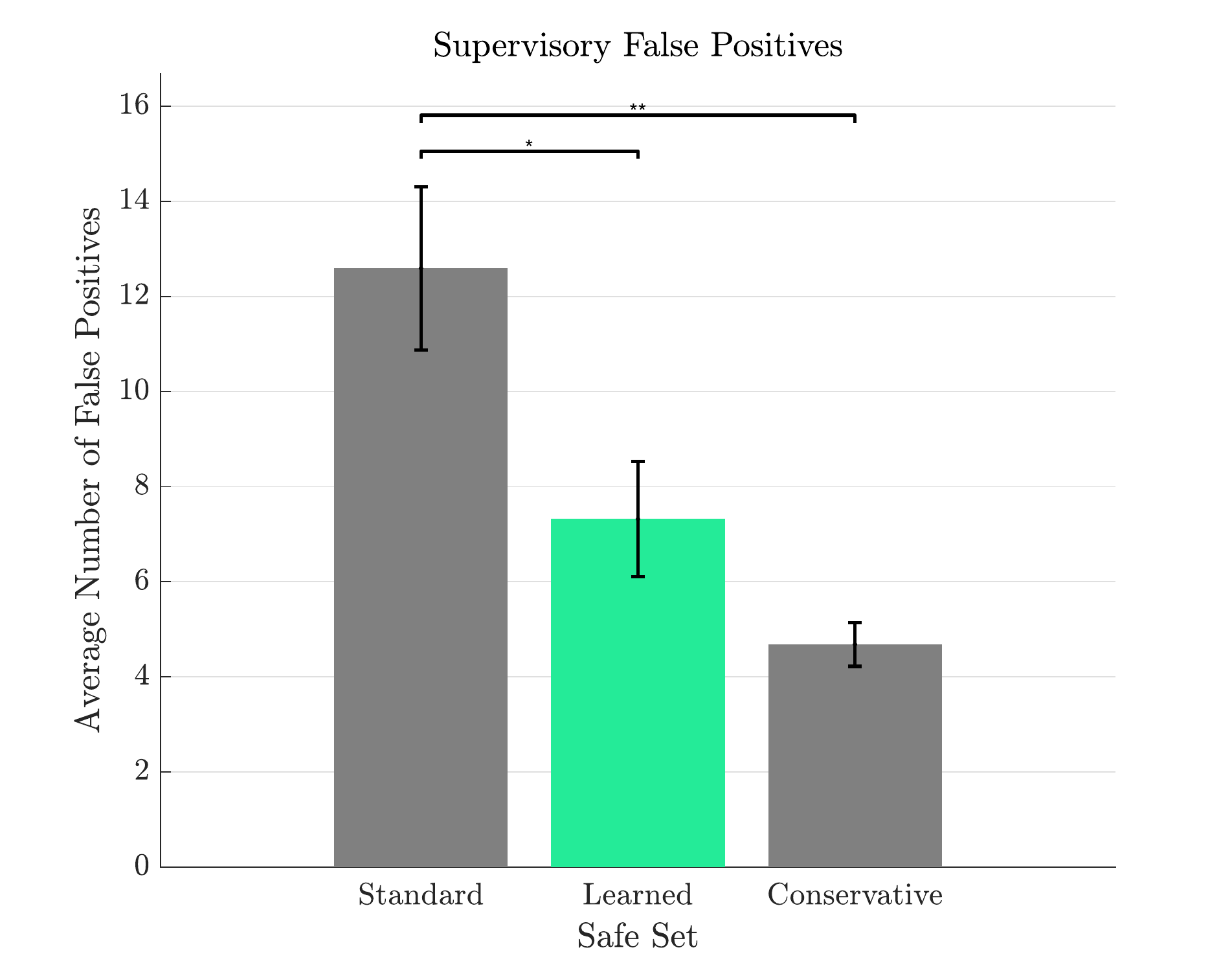}
  \caption{Average number of false positives per trial plotted against the three safe set types.
    There were significant differences between Standard and Learned ($p < .05$) and between Standard and Conservative ($p < .01$).
    There was no significant difference between Learned and Conservative.
  }
  \label{fig:false_positives}
  \vspace{-10pt}
\end{figure}

The second half of that hypothesis, that \textbf{\textit{decreasing supervisory false positives will increase trust and team performance}} was not shown conclusively from our data.
We performed a one-way, repeated measures ANOVA on the pairwise comparison surveys between the teams using the Learned and the Standard safe sets.
Measures of trust showed no significant improvement ($F(1,9) = 1.86$, $p = 0.21$).

\subsection{H2: Preference over Conservative}
For 9 of 11 participants, the Learned safe set had shorter avoidance arcs than the Conservative set.
We hypothesized that this greater efficiency would make the tailored conservativeness of the Learned set preferable to the baseline Conservative safe set.
However, a t-test showed that the survey responses for preference were statistically indistinguishable ($p = 0.8$) from a neutral score: an inconclusive result for Hypothesis 2.
We believe that this result stems from users judging preference more on intelligibility, the ease of avoiding false positives, than on efficiency, the shortness of paths.
As discussed in Section \ref{sec:analysisH1}, both the Learned and Conservative safe sets led to significant false positive reductions over the Standard set.

\begin{figure}[t]
  \vspace{10pt}
  \centering
  \includegraphics[width=\columnwidth]{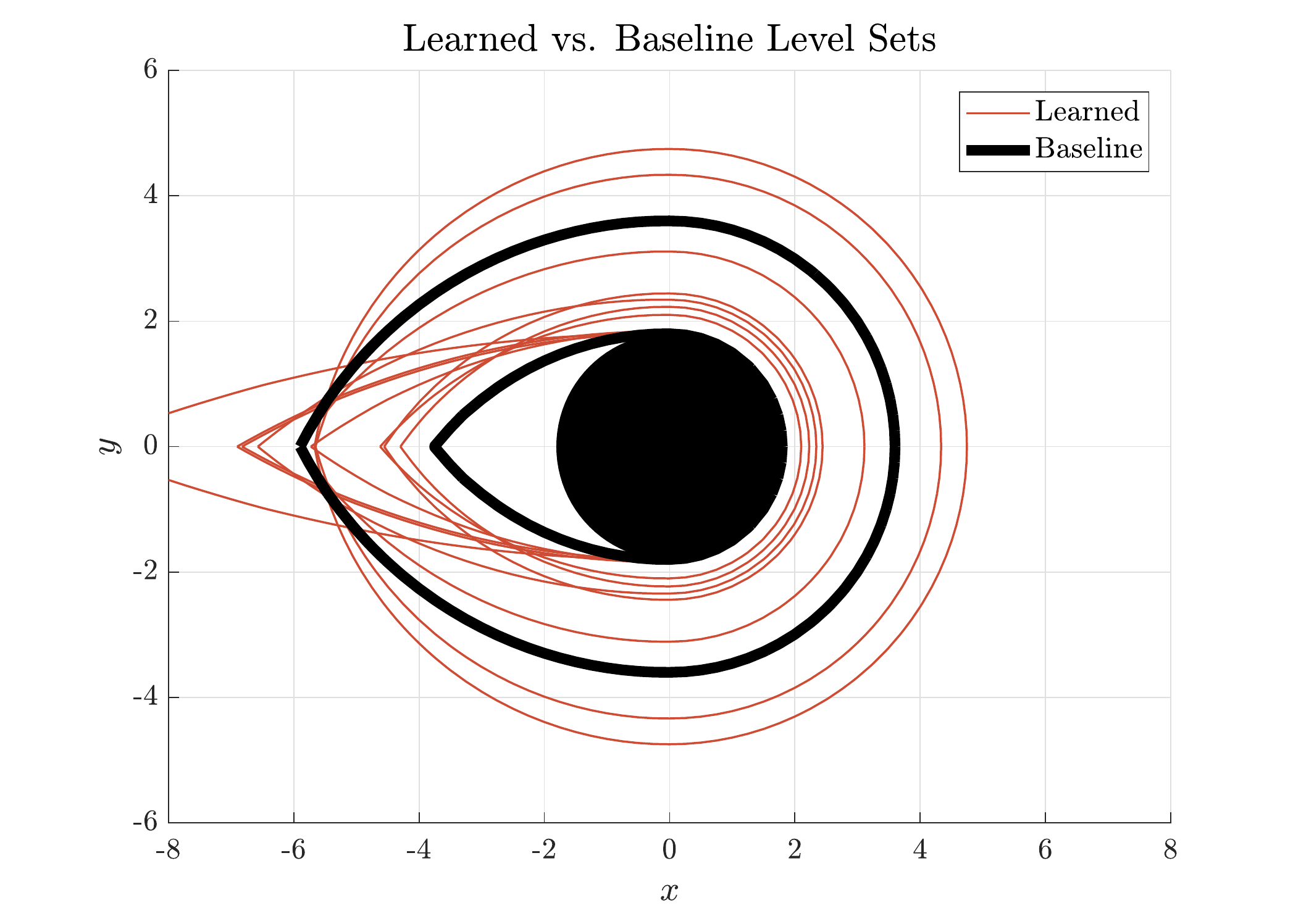}
  \caption{Regressed safe sets (viewed on the $\theta = 0$ slice) from supervisor intervention data overlaid on baselines.
    Three users' safe sets clustered to arcing like the Standard safe set.
    Three others clustered to arcing like the Conservative safe set.
    The final five safe sets exhibit a distinct behavior that reflects supervisors' preference for gradual, pre-emptive arcs.
  }
  \label{fig:learned_safesets}
  \vspace{-10pt}
\end{figure}

This indistinguishability is further compounded since a preference for intelligibility seems to be expressed by some subjects in their Phase II intervention data, resulting in their Learned safe sets having similar arcs as the Conservative safe set (see Fig. \ref{fig:learned_safesets}).
Future work could investigate this efficiency-intelligibility trade-off further by using a conservative baseline that is distinguishably more conservative than user safe sets and by making efficiency more central to the team task.


\subsection{Model Validity}
\label{sec:model_validity}
\begin{figure}[t]
  \centering
  \includegraphics[width=\columnwidth]{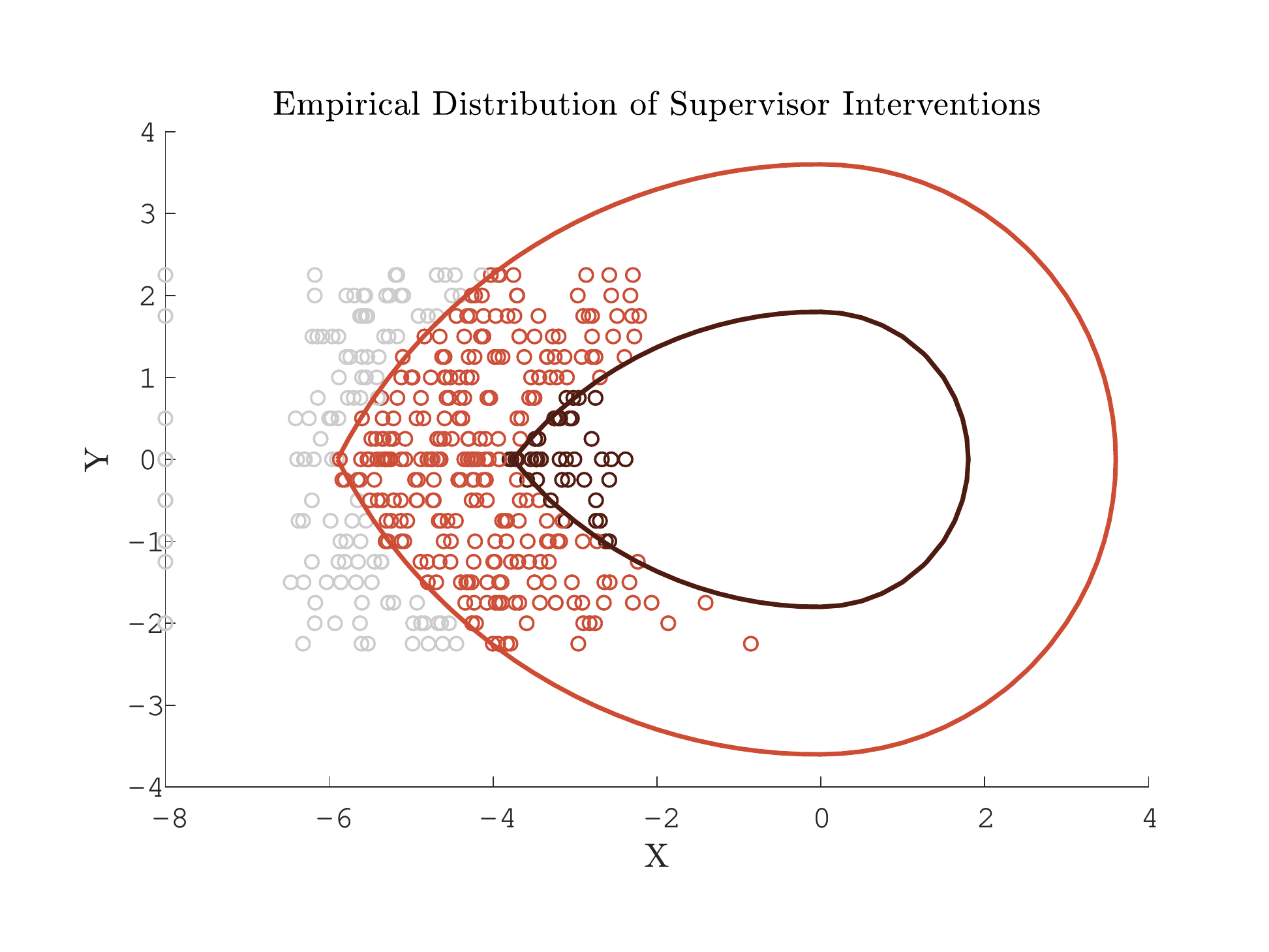}
  \caption{Empirical distribution of intervention states observed during data collection (Phase II of the experiment).
    The interventions within the Conservative reachable set are colored in red, leaving 115 interventions in the corresponding safe set.
    Similarly, the interventions within the Standard reachable set are colored darker, leaving 397 interventions in the corresponding safe set.
    Intervention states not contained within a reachable set would have generated a false positive during the human-robot teaming task.
  }
  \label{fig:validity}
\end{figure}

The statistically significant decreases in false positives observed in Phase III agree with the decreases predicted by the supervisor model based on intervention data from Phase~II.
Our model posits that interventions occur at states noisily distributed about a safe set boundary.
Therefore, it predicts that the empirical distribution of Phase II intervention states contained within a proposed safe set (see Fig. \ref{fig:validity}) will mirror the proportion of false positive interventions observed in Phase III: if states are deemed safe by the controller, they will not be avoided, even when the noisy supervisor would judge them to be unsafe.
Since the Learned safe set controller intervenes at the $\hat{\mu}^*$ level set (see Section \ref{sec:learn_safe_sets}), exactly half the intervention states will be contained within the Learned safe set in expectation.
The model's predictions are compared against observed false positives in Table 1.

\begin{table}
  \centering
  \begin{tabular}{| l | c c | c c |}
    \hline
     & Interventions & Predicted & Average & Observed \\
     & in Safe Set & F.P. vs Std. & F.P. & F.P. vs Std. \\
     \hline
    Standard & 397 / 440 & 100\% & 12.54 & 100\% \\
    Learned & 220 / 440  & 55.4\% & 7.31 & 58.3\% \\
    Conservative & 115 / 440 & 29\% & 4.68 & 37.3\% \\
    \hline
  \end{tabular}
  \vspace{10pt} \\
  Table 1: Predicted and observed false positives.
  Left: Predicted false positives from Phase II data.
  Right: Observed false positives in Phase III.
  \vspace{-10pt}
\end{table}

\section{Conclusion}
\label{sec:conclusion}
Automation with human supervisors relies on leveraging the human supervisor's cognitive resources for success.
Respecting these resources is essential for creating well performing human-robot teams.
It is especially important to avoid overtaxing the human as automated teams continue to scale up, and a single human worker both accomplishes more and bears more cognitive load than ever.
To alleviate this burden, we can decrease the number of issues that command the supervisor's attention by reducing false positives.
By modeling which system states command supervisory attention, we can program autonomous systems to avoid those states when they do not require attention.
To capture this information, we combine the concept of mental simulation from cognitive science with formal safety analysis from reachability theory to propose the noisy idealized supervisor model.
We employ the noisy idealized supervisor as the generative model in a learning algorithm to predict supervisor safety judgements, and we present a safety controller for robotic agents that respects the supervisor's perception of safety.
This safety controller is guaranteed to reduce false positives for idealized supervisors.
Furthermore, for actual supervisors, our human-robot teaming user study demonstrated a significant reduction in false positives when using our approach compared to the standard baseline.

Our results show that it is possible to reduce false positives, and thus cognitive load, by aligning robot behavior with humans' expectations.
Our approach is applicable whenever reachability theory can tractably analyze a dynamical system that will be subject to human safety judgements.
Future work will explore the impact of this framework on application domains from air traffic management to self-driving vehicles.

\addtolength{\textheight}{-12cm}   




%


\bibliographystyle{plain}
\bibliography{PCARI-HI}

\end{document}